\documentclass{article}
\usepackage{ijcai19}

\usepackage{times}
\usepackage{soul}
\usepackage{url}
\usepackage[hidelinks]{hyperref}
\usepackage[utf8]{inputenc}
\usepackage[small]{caption}
\usepackage{graphicx}
\usepackage{amsmath}
\usepackage{booktabs}
\usepackage{algorithm}
\usepackage{algorithmic}
\usepackage{multirow}
\usepackage{booktabs} 
\usepackage{enumerate}
\usepackage{subfigure}
\usepackage{epstopdf}
\usepackage{balance}
\usepackage{flushend}
\usepackage{multirow}
\usepackage{enumerate}
\usepackage{caption}
\usepackage{verbatim}
\usepackage{mathrsfs}
\usepackage{color}

\title{Graph Convolutional Neural Networks via Motif-based Attention}
\author{
Hao Peng$^{1,2}$,
Jianxin Li$^{1,2}$,
Qiran Gong$^{2}$,
Senzhang Wang$^3$,
Yuanxing Ning$^{2}$,
Philip S. Yu$^4$
\affiliations
$^1$ Beijing Advanced Innovation Center for Big Data and Brain Computing, Beihang University\\
$^2$ State Key Laboratory of Software Development Environment, Beihang University\\
$^3$ Collage of Computer Science and Technology, Nanjing University of Aeronautics and Astronautics\\
$^4$ Department of Computer Science, University of Illinois at Chicago\\
\emails
\{penghao, lijx, gongqr, ningyx\}@act.buaa.edu.cn, szwang@nuaa.edu.cn, psyu@uic.edu
}

\begin{document}
\maketitle

\begin{abstract}
Many real-world problems can be represented as graph-based learning problems. In this paper, we propose a novel framework for learning spatial and attentional convolution neural networks on arbitrary graphs. Different from previous convolutional neural networks on graphs, we first design a motif-matching guided subgraph normalization method to capture neighborhood information. Then we implement subgraph-level self-attentional layers to learn different importances from different subgraphs to solve graph classification problems. Analogous to image-based attentional convolution networks that operate on locally connected and weighted regions of the input, we also extend graph normalization from one-dimensional node sequence to two-dimensional node grid by leveraging motif-matching, and design self-attentional layers without requiring any kinds of cost depending on prior knowledge of the graph structure. Our results on both bioinformatics and social network datasets show that we can significantly improve graph classification benchmarks over traditional graph kernel and existing deep models.
\end{abstract}

\section{Introduction}\label{sec:intro}
Graph classification, which aims to identify the class labels of graphs in a dataset, is critically important to many real-world applications in a diverse set of fields.
Data from molecular chemistry and bioinformatics drug discovery ~\cite{Benk2003A}, social network analysis~\cite{backstrom2011supervised}, text classification~\cite{peng2018large}, etc, can all be represented as labeled graphs with relationships and interdependencies between objects.
In chemistry and bioinformatics drug analysis, for instance, each chemical compound can be represented as a graph where nodes correspond to atoms, and edges signify the presence of chemical bonds between atoms.
The task then is to predict the class label of each graph, for instance, the anti-cancer activity, mutagenic or toxicity of a chemical compound.

To solve this problem, one usually extracts certain graph features that help discriminate between graphs of different classes.
Traditional technologies include random-walks, subgraphs or sub-tree patterns based graph kernel methods~\cite{prvzulj2007biological,vishwanathan2010graph,Shervashidze2011Weisfeiler}.
The key idea of the popular random-walk method is to decompose a graph into node paths through random walk and count the co-occurrence of paths on each graph.
Generally speaking, graphs that share a lot of common graphlets are considered similar.
The graph kernel based methods measure the similarity between two graphs with kernel functions corresponding to the inner products of the extracted features~\cite{vishwanathan2010graph,Yanardag2015Deep}.
With the recent success of deep learning techniques, the focus of graph classification techniques has shifted from diverse graph kernel functions to the spatial-based graph convolutional network (GCN) models~\cite{zhou2018graph}.
One pioneering spatial-based graph convolutional neural network model is PATCHY-SAN (PSCN)~\cite{Niepert2016Learning}.
It utilizes standard convolutional neural network (CNN) in GCNs by converting graph-structured data into grid-structured data with sorting functions.
However, PSCN only samples a few of vertex arrangements as the grid-structured data to represent a graph, which may lose a lot of structural information such as the subgraph structures. 
In addition, the design of the sorting functions requires strong prior knowledge, which is difficult in practice.

It is non-trivial to obtain a desirable performance for graph classification due to the following two major challenges.
First, the complexity of graph data has posed significant challenges on existing machine learning algorithms.
Since graph data are in the non-Euclidean domain, each graph has a variable size of unordered nodes and each node in a graph has a different number of neighbors, causing some important operations (e.g., convolutions, recurrences), which are easy to compute in the image/natural language domains, very hard to conduct in the graph data.
Although some recent works including DGCNN~\cite{zhang2018end}, NEST~\cite{Carl2018} and DIFFPOOL~\cite{Ying2018Hierarchical} proposed new graph preprocessing paradigms or graph convolution networks, they cannot fully capture aggregated features from neighbors and nodes by using variety of pooling operators.
Second, exiting deep learning based graph classification models~\cite{zhou2018graph} lack of sufficient study on the diverse impacts of different nodes, graphlets or subgraphs on graph classification due to the difficulty in modeling the impacts between two feature maps generated by convolution kernels. 
The traditional attention model requires strong prior knowledge, which is challenging in arbitrary graph classifications.
Although GAM~\cite{lee2018graph} is a node-level attentional RNN model which considers local connectivity among nodes, we argue that higher level such as subgraph-level attention is more interpretable and important in graph classification.

To address the above challenges, we propose a novel \underline{M}otif-based \underline{A}ttentional \underline{G}raph \underline{C}onvolutional \underline{N}eural \underline{N}etwork model namely MA-GCNN for graph classification. 
We first propose a motif-matching based subgraph normalization method to better capture spatial information as fully as possible by converting graph-structured data into a new grid-structured representation.
Second, we design subgraph-independent convolutional neural networks to learn different-levels of features for each subgraph without pooling operators.
A novel subgraph-level self-attention mechanism is also proposed in the propagation step to learn different impacts or weights to different subgraphs of a graph for graph classification.
We first concatenate the feature maps of different convolution kernels in a subgraph into a vector, and then measure different impacts or weights for each subgraph vector by the self-attention mechanism.
By leveraging the motif-matching based graph processing, subgraph-independent convolution operators and self-attentional layers, we finally design a novel end-to-end graph classification framework.

We conduct extensive experiments on both bioinformatics and social network datasets for graph classification tasks.
Compared to state-of-the-art methods, including traditional graph kernel based algorithms and existing popular deep learning approaches, our proposed models achieve significant improvement in classification accuracy on both two benchmark datasets.

\section{Related Work}\label{sec:relatedwork}
Existing works for graph classification can be broadly categorized into traditional graph kernel based methods and graph convolution neural networks based models.

A great deal of research works have focused on designing the suitable kernel functions for each graph dataset in terms of classification tasks.
Popular methods include graphlets~\cite{Shervashidze2009Efficient}, random walk and shortest path kernel~\cite{Borgwardt2005Shortest}, Weisfeiler-Lehman subtree kernel~\cite{Shervashidze2011Weisfeiler}, deep graph kernel~\cite{Yanardag2015Deep}, graph invariant kernel~\cite{Orsini2015Graph} and multiscale laplacian graph kernel~\cite{Kondor2016The}.
The graphlet kernel decomposes a graph into graphlets, Weisfeiler-Lehman kernel decomposes a graph into subtrees, and Shortest-Path kernel decomposes a graph into shortest-paths.
The decomposed sub-structures are then represented as a vector of frequencies where each item of the vector represents how many times a given sub-structure occurs in the graph.
Thus, the Euclidean space or some other domain-specific reproducing kernel Hilbert space is used to define the dot product between the vectors of frequencies.
In sum, kernel based models can capture sub-structural similarity at different levels, but lack of ability to capture implicit similarities.

Recently, graph convolution neural networks are proposed and presented promising performance in graph classification.
The original idea of defining graph convolution has been recognized as the problem of learning filter parameters that appear in the graph fourier transform in the form of a graph Laplacian~\cite{Bruna2013Spectral}.
~\cite{Kipf2016Semi} proposed a self-loop graph adjacency matrix and a propagation rule to compute and update each neural network layer weights.
An optimized GCNNs model is proposed in~\cite{Defferrard2016Convolutional} by utilizing fast localized spectral filters and efficient pooling operations.
Considering the weakness of traditional CNNs in spatial hierarchies and rotational invariance, a Graph Capsule networks model is proposed in ~\cite{Verma2018Graph}.
~\cite{lee2018graph} proposed a RNN based node-level attention model GAM to processes informative parts of a graph by adaptively visiting a sequence of important nodes.
Due to the disadvantage of partial observability of the input graphs, GAM cannot achieve state-of-art performance in experimental results (not suitable).
Other RNN autoencoders based graph representation methods adopted random walks, breadth-first search and shortest paths to generate node sequences to learn structural features~\cite{Taheri2018}.
~\cite{Simonovsky2017Dynamic} introduced a edge-conditioned convolution operation on graph signal performed in the spatial domain where filter weights are conditioned on edge labels and dynamically generated for each specific input sample.
~\cite{Ying2018Hierarchical} proposed a differentiable graph pooling to generate hierarchical representations of graphs.
In ~\cite{Ivanov2018Anonymous}, authors used distribution of anonymous walks as a network embedding, sampling walks in a graph to approximate actual distribution with a given confidence.
In DGCNN~\cite{zhang2018end}, authors proposed a SortPooling layer which sorts graph vertices in a consistent order so that traditional neural networks can be trained on the graphs.
Different from the above methods to capture spatial structures through various pooling and convolution operations, our proposed models preserves spatial structures by semantically rich grid-structured representation.
Motifs are high-order structures that are crucial in many domains such as bioinformatics, neuroscience and social networks.
Recent work has explored motifs in clustering~\cite{Benson2016Higher} and graph classification~\cite{Monti2018MotifNet,Carl2018} tasks.

However, above-mentioned techniques do not fully exploit motifs to capture local stationary and spatial structures of graph, and they focus on applying motifs to filter structures of dataset to optimize neural networks.
The most relevant work to ours is PSCN~\cite{Niepert2016Learning}, as a standardized process from graph to convolutional neural networks, and PSCN consists of node sequence selection, graph normalization and shallow convolution.
Compared to the proposed models, PSCN model loses lots of structural information.

\section{Motif-matching guided Graph Processing}\label{sec:multiorder}
\begin{figure}[t]
\centering
\includegraphics[width=0.45\textwidth]{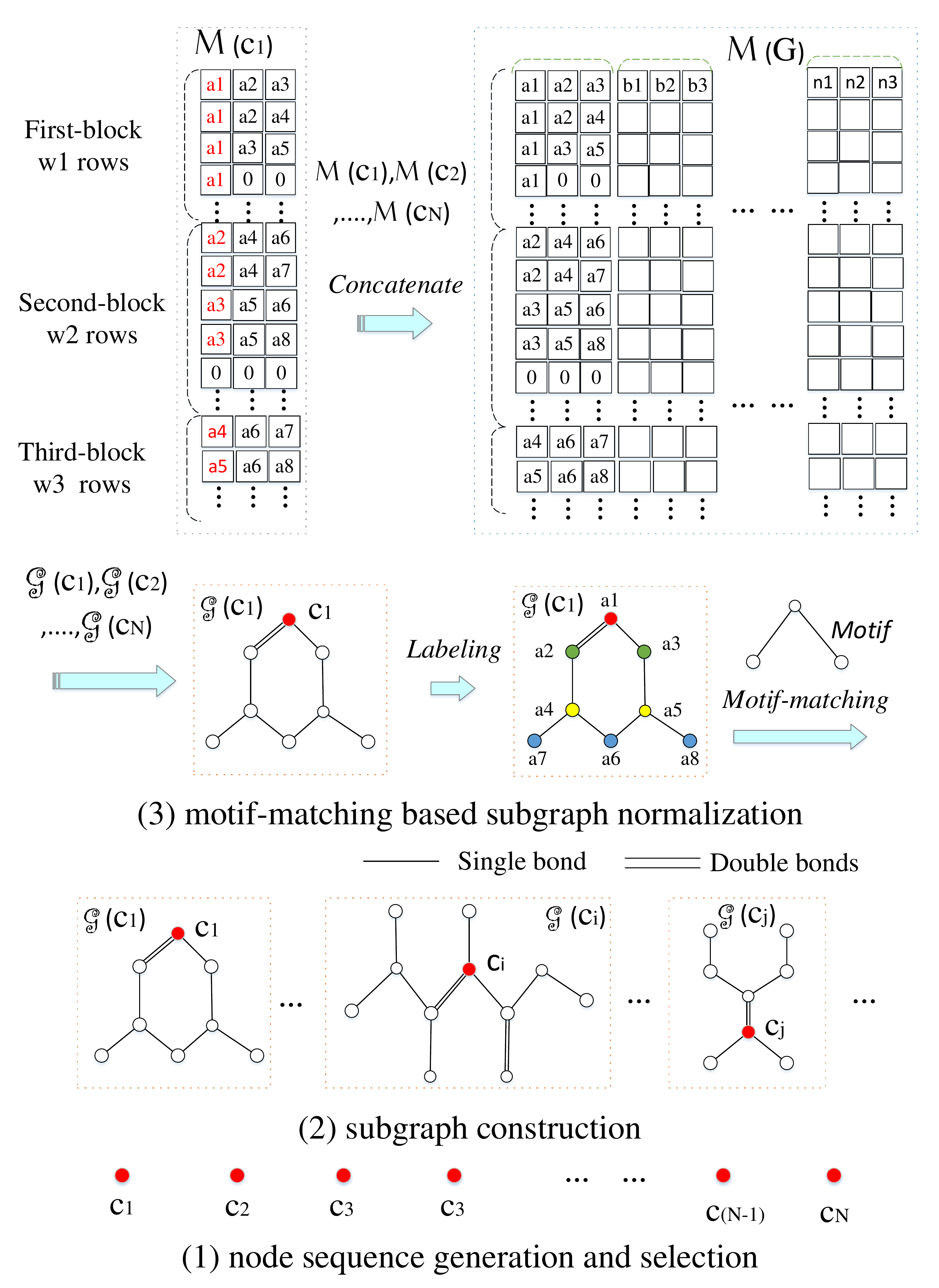}\vspace{-0.15in}
\caption{An illustration of motif-matching guided graph processing. 
In real datasets, there are some edges with double or triple bonds, and we treat it as edge weights.}\label{fig:data_processing}\vspace{-0.15in}
\end{figure}

In this section, we introduce the motif-matching guided graph processing to transform the graph-structured data to grid-structured representation that can preserve the rich semantic information.
The proposed novel graph processing including node sequence generation and selection, subgraph construction and motif-matching based subgraph normalization, as shown in Figure~\ref{fig:data_processing}.

In our models, we use the two-hop paths motif as the motif structure, because two-hop paths motif has symmetrical properties and is very suitable for local matching.
The distributions of edges in graphs-structured data are usually unbalanced in general, so some triangles or more complex motifs are rarely matched.
Even, dense subgraphs can be partitioned into multiple two-hop paths motifs and can be approximately reconstructed.
The two-hop paths motif can be easily represented by a array which is suitable for convolution operations.

Second, we denote an arbitrarily graph as $G = (V, E)$ with $n$ vertices and $m$ edges, and $d(v, u)$ denotes the shortest-path between any two nodes $v$ and $u$ in $G$.
We can compute the \emph{closeness centrality} $C_{v} ={(n-1)}/{\sum_{u\in{V}, u\neq v}d(v,u)}$ for each node $v$ in the graph $G$.
Here, we treat the double, triple and higher bonds of the edges in bioinformatics as two, three and more single bonds in graph.
For the node sequence generation and selection, we sort all nodes in a graph by their \emph{closeness centrality}, and then select the top-$N$ nodes in the sequence as central nodes of the graph.
As shown in the step (1) of Figure~\ref{fig:data_processing}, the red nodes from $c_1$ to $c_N$ represent the selected central nodes.
Then, we extract a subgraph $\mathcal{G}(c_i)$ for each central node $c_i, i\in[1,N]$, as shown in step (2) of Figure~\ref{fig:data_processing}.
Here, we limit the number of nodes in the subgraph to be no more than $K$, and the subgraph is extracted in the order of \emph{first}, \emph{second} and \emph{third} order neighbors using breadth first search (BFS) and their \emph{closeness centrality}.

Third, for each node $\bar{u}$ in subgraph $\mathcal{G}(c_i)$, we calculate a shortest distance between node $\bar{u}$ to the central node $c_i$ within the subgraph.
Then we sort the nodes in the subgraph $\mathcal{G}(c_i)$ by the \emph{closeness centrality} and then the shortest distance to the central node $c_i$.
So, we can label the nodes based on the above sorted index, such as $\{a1, a2, \cdots, a8\}$ in $\mathcal{G}(c_1)$, as shown in the step (3) of Figure~\ref{fig:data_processing}.
Here, we divide the \emph{central matrix} $\mathcal{M}(c_i)$ into three blocks according to the shortest distance from the matched motif to the center node $c_i$, and initialize them with zeros, as shown in the step (3) of Figure~\ref{fig:data_processing}.
Each row of \emph{central matrix} $\mathcal{M}(c_i)$ is filled with matched nodes in the two-hop paths motif\footnote{As the number of nodes in the subgraph is small, we can also perform a fast motif matching algorithm~~\cite{Sun2012Efficient} for large subgraphs.}.
So, the first block contain central node, such as the $a1$ nodes preserved in the first block of $\mathcal{M}(c_1)$ shown in the step (3) of Figure~\ref{fig:data_processing}.
Similarly, the second and third blocks should contain 2-hop and 3-hop nodes from the center node, respectively.
In general, as shown in Figure~\ref{fig:data_processing}, different hops contain different numbers of blocks, such as $\{a1, a2, a3\}, \{a1, a2, a4\}$ and $\{a1, a3, a5\}$ in the first block, $\{a2, a4, a6\}, \{a2, a4, a7\}$, $\{a3, a5, a6\}, \{a3, a5, a8\}$ in the second block and $\{a4, a6, a7\}, \{a5, a7, a8\}$ in the third block of $\mathcal{M}(c_1)$.
We also guarantee that the nodes with smaller sorted indexes in the three nodes are ranked ahead.
To represent the matrices of subgraphs with different scales in a unified way, we fix the row numbers of the first, second and third blocks as $w_{1}, w_{2}$ and $w_{3}$, respectively.
For each subgraph in the entire graph, we generate a corresponding center matrix.
Note that we save all the nodes in the subgraph to the corresponding center matrix according to the above sorting functions.
Next, we concatenate the $N$ \emph{central matrices} $\mathcal{M}(c_i), i\in[1,N]$ into a large matrix $M(G)$ following the sequence in step (1), such as the the combination of $\mathcal{M}(c_1)$, $\mathcal{M}(c_2), \cdots ,\mathcal{M}(c_N)$ shown in step (3) of Figure~\ref{fig:data_processing}. 
So, for an entire graph $G$, it can be represented by a matrix $\mathcal{M}(G)$.
We note that each element in the matrix $\mathcal{M}(G)$ refers to a vertex in the graph $G$.

\begin{figure*}\vspace{-0.2in}
\centering
\includegraphics[width=0.85\textwidth,height=0.3\textheight]{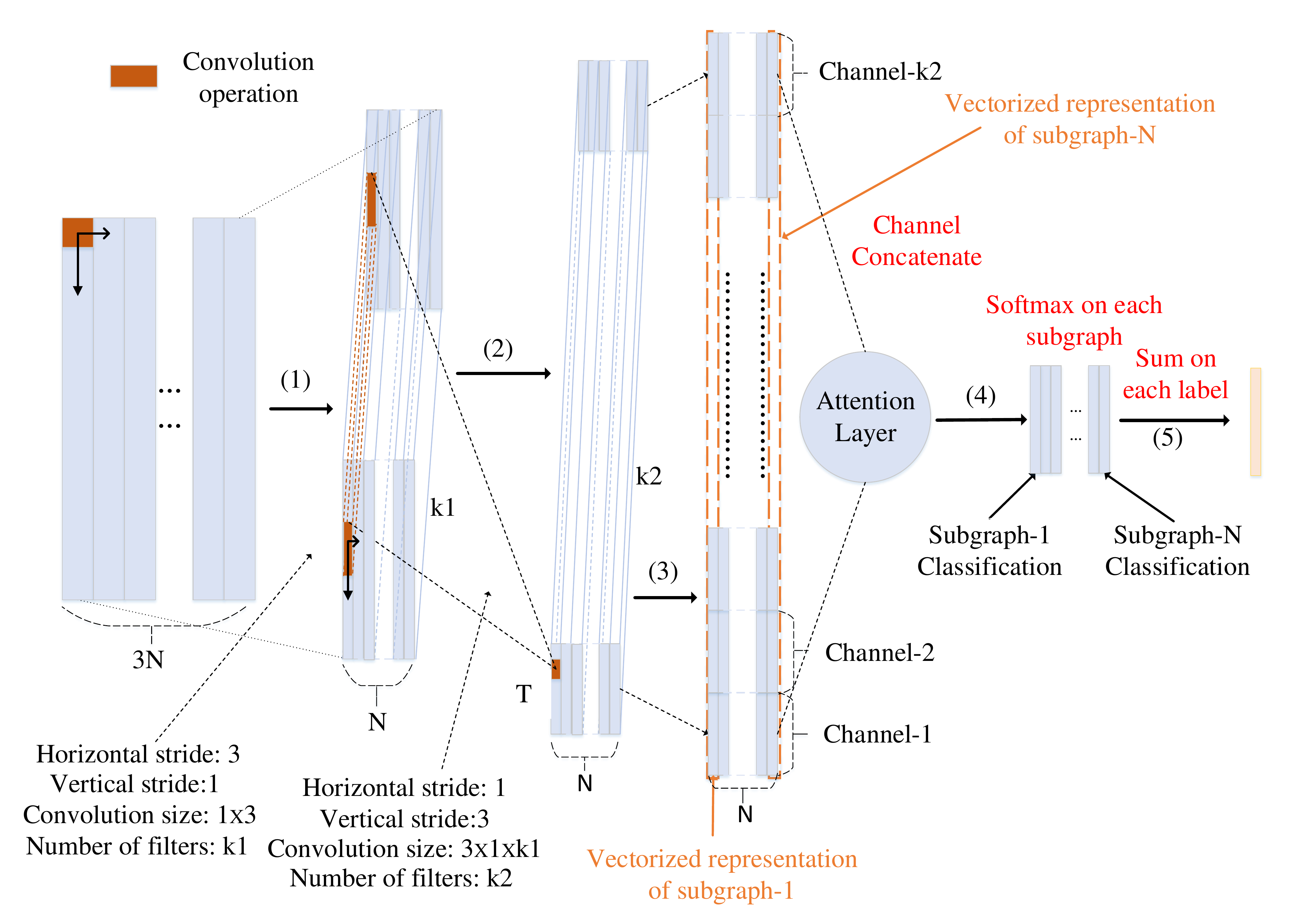}\vspace{-0.2in}
\caption{An illustration of the subgraph-level self-attention deep convolutional neural networks. The numbers of convolutional layers, attention layers and loss function can be adjusted and based on the size of datasets and total labels.}\label{fig:attention_gcn}\vspace{-0.15in}
\end{figure*}

\section{Motif-based Attentional Graph Convolutional Neural Networks}\label{sec:Motif-GCNN}
After transforming arbitrarily graph $G$ to the matrix $\mathcal{M}(G)$, we use convolutional neural networks and attention layer to learn different-levels features of graph.
The size of the input matrix $M(G)$ is $3N\times(w_{1}+w_{2}+w_{3})$, where $N$ is the number of selected central nodes, 3 is the length of the motif, and $(w_{1}+w_{2}+w_{3})$ is the sum of the rows of the three matrix blocks.
In the first convolution layer, the size of the convolution kernel is $3\times1$, and the convolution slides in both horizontal and vertical directions.
In order to ensure the independence of feature extraction between subgraphs, the horizontal direction stride is 3, which equals to the length of the motif, and the vertical direction stride is 1.
We use $K_1$ convolution kernels to generate a $K_1 \times N\times (w_{1}+w_{2}+w_{3})$ features map, where each vertical representation characterizes the extracted features of corresponding subgraph as shown in the step (1) of Figure~\ref{fig:attention_gcn}.
In the second layer, the size of convolution kernel is $3\times1\times K_1$.
To guarantee the independence of subgraphs' feature learning, the horizontal stride is 1 and the vertical stride is 3.
We do not employ any pooling operations to preserve spatial information of the graph.
We denote $T = (w_{1}+w_{2}+w_{3})/3$, and the size of feature map of the second convolution layer is $N\times K_2\times T$ with $K_2$ convolution kernels, as shown in the step (2) of Figure~\ref{fig:attention_gcn}.
So far, we can directly implement the two layers of convolutional neural network to learn different-levels of features for each graph by adding a softmax network for graph classification tasks.
We name the combination of \underline{M}otif-matching guide \underline{G}raph processing, two layers of subgraph-independent \underline{C}onvolutional \underline{N}eural \underline{N}etwork and two layers of fully-connected networks as \textbf{M-GCNN}.

We further study the different influences of each subgraph in the graph classification.
Different from traditional softmax networks, we implement a self-attentional layer to capture the different influences among subgraphs.
In order to characterize the features of each subgraph as a vector, we concatenate each channel of feature map into a $N\times (K_2\times T)$ matrix, where each dimension in $N$ is the representation of corresponding subgraph.
The vectorized representation of the features of each subgraph is shown in the step (3) of Figure~\ref{fig:attention_gcn}.
For any two subgraphs $\mathcal{G}(c_i)$ and $\mathcal{G}(c_j)$, the corresponding features vectors are $\vec{h}_{c_i}$ and $\vec{h}_{c_j}$, where the size of the features vector is denoted as $F = K_2\times T$.
Then, we add a $W \in R^{F' \times F}$ weight matrix and a $\vec{a}^{T}\in R^{2F'}$ weight vector to learn mutual influences among subgraphs.
More precisely, the attentional layer contains $F'$ hidden neurons, and the attention mechanism is a feed forward neural network.
When applying the LeakyReLU (with negative input slope $\alpha = 0.2$), the coefficients of subgraph $\mathcal{G}(c_j)$ on $\mathcal{G}(c_i)$ computed by the attention mechanism can be formalized as:
\begin{equation}\label{eq:attention}
\centering
\alpha_{ij} = \frac{exp(LeakyReLU(\vec{a}^{T}[W\vec{h}_{c_i}\parallel W\vec{h}_{c_j}]))}{\sum_{k\in N, k\neq i }exp(LeakyReLU(\vec{a}^{T}[W\vec{h}_{c_i}\parallel W\vec{h}_{c_k}]))},
\end{equation}
where $\parallel$ denotes the concatenate operation.
Following the attention mechanism, we perform $S$ independent attention computations, and employ \emph{averaging} strategy to evaluate influences.
As shown in the step (4) of Figure~\ref{fig:attention_gcn}, the output nonlinearity uses a softmax for final classification:
\begin{equation}\label{eq:multi-attention}
\centering\vspace{-0.1in}
\vec{h'}_{C_i} = \sigma(\frac{1}{S}\sum^{S}_{s=1}\sum_{t\in N, t\neq i}\alpha^{s}_{it}W^{s}\vec{h}_{C_t}),
\end{equation}
where $\sigma$ is the sigmoid function, and $\vec{h'}_{c_i}$ is the output feature of subgraph $\mathcal{G}(c_i)$.
Finally, we sum the output features according to the class label, as shown in the step (5) of Figure~\ref{fig:attention_gcn}.
The maximum value represents the class it belongs to.
We name this framework as \textbf{MA-GCNN}.

The reason for the subgraph-level self-attention approach is that we consider each subgraph can partially represent the graph. 
Each subgraph contributes differently on the graph classification.
Through the attentional layer, we can obtain the classification on each subgraph. 
Then we can ensemble the classification results of subgraphs by applying sum operation and a softmax layer as marked in red in Figure~\ref{fig:attention_gcn}.

\section{Experiments}\label{sec:exper}
In the experiments, we compare our proposed algorithms with state-of-the-art traditional graph kernels based classification models and recently developed deep learning approaches.

\subsection{Datasets and Settings}
We evaluate our models on two types of real-world graphs including bioinformatics and social network datasets. 
Table~\ref{tab:datasets} summarizes the statistics of the ten datasets.

Bioinformatics datasets contain five categories, namely \textbf{MUTAG}, \textbf{PTC}, \textbf{PROTEINS}, \textbf{D\&D} and \textbf{NCI1}.
\textbf{MUTAG} is a dataset of 188 mutagenic aromatic and heteroaromatic nitro compounds~\cite{Debnath1991Structure} with 7 discrete node labels, namely \emph{C, N, O, F, I, Cl, Br}, and 4 discrete edge labels, including aromatic, single, double, and triple bonds.
The classes indicate whether the compound has a mutagenic effect on a bacterium.
PTC~\cite{Toivonen2003Statistical} is a dataset of 344 organic molecules marked according to their carcinogenicity on male and female mice and rats, and it has 19 discrete labels in nodes.
PROTEINS is a graph collection obtained from~\cite{Borgwardt2005Protein} where nodes are secondary structure elements and edges indicate neighborhood in the amino-acid sequence or in 3D space with 61 discrete labels.
The graphs are classified as enzyme or non-enzyme. 
D\&D is a dataset of 1178 protein structures ~\cite{Dobson2003Distinguishing} with 82 discrete labels, and is also classified into enzymes and non-enzymes. 
NCI1 dataset is chemical compounds screened for activity against non-small cell lung cancer and ovarian cancer cell lines~\cite{Wale2008Comparison}, and contains 4110 samples.
In order to test the effectiveness of our algorithms on unlabeled graphs, we also choose five social network datasets, including \textbf{IMDB-BINARY}, \textbf{IMDB-MULTI}, \textbf{REDDIT-BINARY}, \textbf{REDDIT-MULTI-5K} and \textbf{REDDIT-MULTI-12K}. 
Note that we use node degree as the attribute in these datasets, and it can easily incorporate continuous features.
Both IMDB-BINARY and IMDB-MULTI are movie collaboration datasets, and contain 1000 and 1500 graphs, respectively.
The task is to identify which genre an ego-network graph belongs to. 
REDDIT-BINARY, REDDIT-MULTI-5K and REDDIT-MULTI-12K are datasets where each graph corresponds to an online discussion thread and their nodes correspond to users. 
There is an edge between two nodes if at least one of them responds to the other’s comment.
The task in these datasets is to predict which subreddit a given discussion graph belongs to.

All of our experiments were performed on 64 core Intel Xeon CPU E5-2680 v4@2.40GHz with 512GB RAM and 4$\times$NVIDIA Tesla P100-PICE GPUs. 
The operating system and software platforms are Ubuntu 5.4.0, Tensorflow-gpu (1.4.0), and Python 2.7.
The common parameters of training the models were empirically set, such as MOMENTUM = 0.9, Dropout = 0.5, learning rate = 0.001, $L2$ norm regularization weight decay = 0.01, etc.
We set $F1 = 128, F2 = 64$ in M-GCNN model, and $S = 8$ in MA-GCNN model.
For each dataset, the parameters $N, K, w_{1}, w_{2}, w_{3}$ and epoch are set by the following principle:
1) the value of $N$ is the average number of nodes for each dataset;
2) the numbers of nodes $K$ in the subgraph to be 10 and 20 in bioinformatics and social network datasets; and 
3) the numbers of $ w_{1}, w_{2}, w_{3}$ are given by the sub-graph connectivity information.
Our algorithm is efficient to converge with the epoch numbers varying from 20 to 500 for different datasets.
Considering the number of training sample and downward trend of the objective function, we adjust the batch size from $45$ to $450$ to get the best accuracy.
We employ the cross-entropy loss function which is widely used in classification tasks.
For all these datasets, we randomly sampled 10\% of the graphs as the testing set, while the remaining graphs are used to perform 10-fold cross-validation to train and evaluate the model.
We report the average prediction accuracy and standard deviations.

\begin{table}[t]
    \centering \small
    \caption{Properties of the datasets.\label{tab:datasets}}\vspace{-0.1in}
    \begin{tabular}{ccccccc}
        \hline
        Datasets& Graphs &Classes &Nodes &Edges &Labels\\
         & & (Max) & Avg & Avg & Vertex\\
        \hline
        MUTAG & 188 & 2(125) & 17.93 & 19.79 & 7\\
        PTC & 344 & 2(63) & 14.29 & 14.69 & 19\\
        PROTEINS & 1113 & 2(619) & 39.06 & 72.82 & 61\\  
        D\&D & 1178 & 2(691) &284.32 &  715.65 & 82\\ 
        NCI1 & 4110 & 2(111) &29.87 & 32.30 & 37\\ 
        \hline  
        IMDB-B & 1000 & 2(500) &19.77 & 193.06 & - \\ 
        IMDB-M & 1500 & 3(500) &13 & 131.87 & - \\ 
        RE-B & 2000 & 2(1000) &429.6 & 995.50 & - \\ 
        RE-M-5K & 5000 & 2(1000) &508.5 & 1189.74 & - \\ 
        RE-M-12K & 11929 & 11(2592) &391.4 & 913.78 & - \\ 
        \hline  
    \end{tabular}\vspace{-0.1in}
\end{table}

\subsection{Baseline Methods}
We compare our model with both traditional graph kernel methods and deep learning based graph classification approaches.
Graph kernel based baselines include the \textbf{Graphlet Kernel (GK)}~\cite{Shervashidze2009Efficient}, the \textbf{Shortest-Path Kernel (SP)}~\cite{Borgwardt2005Protein}, \textbf{Weisfeiler-Lehman Sub-tree kernel (WL)}~\cite{Shervashidze2011Weisfeiler} and \textbf{Deep Graph Kernels (DGK)}~\cite{Yanardag2015Deep}.
For deep learning based  approaches, the following eight state-of-the-art GCNs are compared with: \textbf{PATCHY-SAN (PSCN)}~\cite{Niepert2016Learning}, \textbf{Dynamic Edge CNN (ECC)}~\cite{Simonovsky2017Dynamic}, \textbf{Deep Graph Convolution Neural Network (DGCNN)}~\cite{zhang2018end}, \textbf{Graph Capsule CNN (GCAPS-CNN)}~\cite{Verma2018Graph}, \textbf{Anonymous Walk Embeddings (AWE)}~\cite{Ivanov2018Anonymous}, \textbf{Sequence-to-sequence Neighbors-to-node Previous predicted (S2S-N2N-PP)}~\cite{Taheri2018}, \textbf{Network Structural ConvoluTion (NEST)}~\cite{Carl2018} and \textbf{differentiable graph pooling model (DIFFPOOLS)}~\cite{Ying2018Hierarchical}.
DGCNN enhances the pooling network by solving the underlying graph structured tasks. GCAPS-CNN combines the advantages of spectral domain GCN and capsule networks, which further explores the permutation invariant for graph data.
We implement PSCN model based on the paper ~\cite{Niepert2016Learning}.
For other baselines, we follow exactly the same experiment and model settings as mentioned in the corresponding papers.
Parts of results are not presented because either they are not previously reported or the source code is not publically available.

\begin{table*}[t]
    \centering \small
    \caption{Comparison of classification average accuracy and standard deviation on bioinformatics datasets.}\label{tab:bioexp}\vspace{-0.1in}
    \begin{tabular}{cccccccc}  
        \hline  
        Methods& MUTAG & PTC  & PROTEINS & D\&D & NCI1\\
        \hline
        SP & 87.28$\pm$0.55 & 58.24$\pm$2.44 & 75.07$\pm$0.54 & 78.45$\pm$ 0.26 & 73.47$\pm$0.11 \\
        WL & 83.78$\pm$1.46 & 57.97$\pm$0.49 &  74.68 $\pm$0.49& 79.78$\pm$0.36 & \textbf{84.55$\pm$0.36}  \\
        GK & 81.66$\pm$2.11 & 57.26$\pm$1.41 & 71.67$\pm$0.55 & 78.45$\pm$0.26 & 62.28$\pm$0.29 \\
        DGK & 87.44$\pm$2.72 & 60.08$\pm$2.55 &  75.68$\pm$0.54 & 78.50$\pm$0.22& 80.31$\pm$0.46 \\  
         \hline 
        PSCN & \textbf{92.63$\pm$4.21} & 62.29$\pm$5.68 & 75.89$\pm$ 2.76 & 77.12$\pm$2.41 & 78.59$\pm$1.89 \\
        ECC & 89.44 & - & - & 74.10 & \textbf{83.80} \\ 
        DGCNN & 85.83$\pm$1.66 & 58.59$\pm$2.47 &  75.54$\pm$0.94 & 79.37$\pm$0.94 & 74.44$\pm$0.47 \\
        GCAPS-CNN & - & 66.01$\pm$5.91 & 76.40$\pm$4.17 & 77.62$\pm$4.99 & 82.72$\pm$2.38 \\ 
        AWE & 87.87$\pm$9.76 & - & - & 71.51$\pm$4.02 & - \\ 
        S2S-N2N-PP & 89.86$\pm$1.1 & 64.54$\pm$1.1 & 76.61$\pm$0.5 & - &  \textbf{83.72$\pm$0.4} \\ 
        NEST & 91.85$\pm$1.57 & \textbf{67.42$\pm$1.83} & 76.54$\pm$0.26 & 78.11$\pm$0.36& 81.59$\pm$0.46 \\
        DIFFPOOL & - & - & \textbf{78.10} & \textbf{81.15} & - \\
        \hline
        M-GCNN & \textbf{92.78$\pm$3.56} & \textbf{70.30$\pm$3.59} & \textbf{78.19$\pm$1.93} & \textbf{81.37$\pm$1.11} & 80.91$\pm$2.17 \\
        MA-GCNN & \textbf{93.91$\pm$2.95} & \textbf{71.77$\pm$2.13} & \textbf{79.35$\pm$1.74} & \textbf{81.48$\pm$1.03} & 81.77$\pm$2.36 \\
        \hline 
        Gain & 1.28 & 4.35 & 1.25 & 0.33 & - \\
        \hline  
    \end{tabular}\vspace{-0.1in}
\end{table*}

\begin{table*}[t]
    \centering\small
    \caption{Comparison of classification average accuracy and standard deviation on social network datasets.}\label{tab:socialexp}\vspace{-0.1in}
    \begin{tabular}{ccccccc}
        \hline  
        Methods& IMDB-B & IMDB-M & RE-BINARY & RE-MULTU-5K & RE-MULTU-12K \\
        \hline
        WL &  73.40$\pm$4.63 & 49.33$\pm$4.75 & 81.10$\pm$1.90& 49.44$\pm$2.36 & 38.18$\pm$1.30\\
        GK &  65.87$\pm$0.98 & 43.89$\pm$0.38 & 77.34$\pm$0.18& 41.01$\pm$0.17& 31.82$\pm$0.08\\
        DGK & 66.96$\pm$0.56 &44.55$\pm$0.52 & 78.04$\pm$0.39 & 41.27$\pm$0.18& 32.22$\pm$0.10\\
        \hline
        PSCN & 71.00$\pm$2.29 & 45.23$\pm$2.84& 86.30$\pm$1.58& 49.10$\pm$0.70 & 41.32$\pm$0.32\\
        DGCNN & 70.03$\pm$0.86 & 47.83$\pm$0.85& 76.02$\pm$1.73 & 48.70$\pm$4.54& - \\
        GCAPS-CNN &  71.69$\pm$3.40 & 48.50$\pm$4.10 & 87.61$\pm$2.51 & 50.10$\pm$1.72 & -\\
        AWE &  \textbf{74.45$\pm$5.83} &  51.58$\pm$4.66 &  87.89$\pm$2.53 & \textbf{54.74$\pm$2.93} & 41.51$\pm$1.98\\
        S2S-N2N-PP &  73.8$\pm$0.7 & 51.19$\pm$0.5 & 86.50$\pm$0.8 & 52.28$\pm$0.5 & 42.47$\pm$0.1 \\ 
        NEST & 73.26$\pm$0.72 & \textbf{53.08$\pm$0.31} & \textbf{88.52$\pm$0.64} & 48.61$\pm$0.46 & 42.80$\pm$0.28 \\
        DIFFPOOL & - & - & - & - & \textbf{47.04} \\
        \hline 
        M-GCNN & \textbf{75.10$\pm$3.14} &  \textbf{52.19$\pm$2.66} & \textbf{88.06$\pm$1.29} & \textbf{55.62$\pm$2.19} & \textbf{47.35$\pm$1.31} \\
        MA-GCNN & \textbf{77.20$\pm$2.96} &  \textbf{53.77$\pm$3.11} & \textbf{89.44$\pm$1.18} & \textbf{56.18$\pm$1.48} & \textbf{48.14$\pm$1.93}\\
        \hline 
        Gain & 2.75 & 0.69 & 0.92 & 1.44 &  1.10\\
        \hline  
    \end{tabular}\vspace{-0.1in}
\end{table*} 

\subsection{Experiment Results}
\textbf{Bioinformatics Graph Classification}.
Table~\ref{tab:bioexp} shows the accuracy and standard deviations of different algorithms on the five bioinformatics datasets.
We mark the top-3 scores in bold.
One can see that M-GCNN achieves high accuracy and low standard variances for all bioinformatics datasets.
Even though only the motif-matching guided graph processing is utilized, M-GCNN surpasses all the baseline methods in terms of accuracy with only one exception on NCI1, where the best performing method, WL, does not perform well on most other datasets, especially on PTC and MUTAG.
The performance improvement shows that the proposed motif-matching based subgraph normalization method successfully and fully preserves spatial information in converting graph-structured data into grid-structured data.
Confidently, our proposed subgraph-level self-attention optimized model MA-GCNN achieves the highest accuracy on four datasets.
Compared with M-GCN, MA-GCNN achieves performance improvements in terms of average accuracy and standard variance for all the datasets.
Even on the MUTAG, four out of ten cross-validations, the accuracy rates are 100\%, and the average of accuracy is 93.91\%.
For PTC, MA-GCNN also achieves the highest accuracy 71.77\%, and improves the best baseline NEST by 4.35\%.
MA-GCNN achieves up to 1.25\% and 0.33\% performance improvements over DIFFPOOL model with smaller standard deviations on PROTEINS and D\&D datasets.
For NCI1, MA-GCNN achieves state-of-the-art result on six out of ten benchmarks.
Compared to the most relevant PSCN model, MA-GCNN achieves 3.18\% improvements in terms of average accuracy in NCI1.

\textbf{Social Network Graph Classification}.
Table~\ref{tab:socialexp} shows the results of different algorithms on the five social network datasets.
We employ the normalized node degree as node attribute for social network datasets.
The nodes of the social data have richer node attributes than the bioinformatics data, and have continuity characteristics.
We also mark the top-3 scores in bold.
One can see that both M-GCNN and MA-GCNN models achieve higher accuracy compared with most deep learning based approaches and graph kernel based models. 
The MA-GCNN model achieves the best results on all the five datasets.
In addition, MA-GCNN outperforms M-GCN in all the cases.
For IMDB-B and IMDB-M datasets, MA-GCNN achieves the highest accuracy up to 77.20\% and 53.77\%, respectively.
For REDDIT-BINARY (RE-BINARY), REDDIT-MULTU-5K (RE-MULTU-5K) and REDDIT-MULTU-12K (RE-MULTU-12K) datasets, MA-GCNN achieves the highest accuracy up to 89.44\%, 56.18\% and 48.14\%, respectively. 
MA-GCNN achieves 2.75\%, 0.69\%, 0.92\%, 1.44\% and 1.10\% performance improvements in terms of average accuracy compared with the best baselines, with small standard deviations on the five datasets, respectively.

In summary, M-GCNN and MA-GCNN models show much more promising results against both the recently developed state-of-art deep learning approaches and graph kernels methods.
The improvements in both bioinformatics datasets and social network datasets demonstrate the effectiveness of the motif-matching based subgraph normalization method and the subgraph-level self-attention mechanism.

\section{Conclusion}\label{sec:conclu}
In this paper, we propose a novel motif-matching based subgraph normalization method to preserve spatial information of subgraph.
By integrating motif-matching based graph processing, subgraph-independent convolutional networks and subgraph-level self-attention layer, the proposed MA-GCNN model is able to learn more discriminative and richer features for real-world graph datasets in classification tasks. Extensive evaluations show that MA-GCNN achieves new state-of-the-art performance in both bioinformatics datasets and social network datasets.
In the future, we plan to further study the interpretability of the subgraph-level self-attention GCNN by analysing the distributions of the weights of subgraphs for graph classification.

\clearpage
\bibliographystyle{named}
\bibliography{ms}

\end{document}